\documentclass[10pt,oneside,onecolumn,aps,pra,preprintnumbers,bibnotes,superscriptaddress]{revtex4-2}

\usepackage[utf8]{inputenc}
\usepackage{graphicx}
\usepackage{subcaption}
 
\usepackage{amsmath}
\usepackage{amssymb}
\usepackage{amsfonts}
\usepackage{amsthm}
\usepackage{mathtools}
\usepackage[separate-uncertainty=true]{siunitx}
\usepackage{isomath}
\usepackage{bm}
\usepackage{physics}
\usepackage{cancel}

\usepackage[dvipsnames]{xcolor}
\usepackage{hyperref}
\usepackage{comment}
\usepackage{tikz}
\usetikzlibrary{decorations.pathmorphing}

\newcommand{\mus}{\mu_\text{s}}
\newcommand{\musr}{\mu_\text{s}^\prime}
\newcommand{\mua}{\mu_\text{a}}
\newcommand{\musu}{\mu_{\text{s},\uparrow}^\prime}
\newcommand{\muau}{\mu_{\text{a},\uparrow}}
\newcommand{\musd}{\mu_{\text{s},\downarrow}^\prime}
\newcommand{\muad}{\mu_{\text{a},\downarrow}}

\begin{document}

\title{Learning-Based Reconstruction of Optical Properties in Bilayered Media from Single-distance Time-Resolved Reflectance Measurements}

\author{Caterina Amendola}\thanks{Equally contributing}
\affiliation{Department of Physics, Politecnico di Milano, Milan, 20133, Italy}
\author{Lorenzo Buffoni}\thanks{Equally contributing}
\affiliation{Department of Physics and Astronomy, University of Florence, Sesto Fiorentino, 50019, Italy}
\author{Giulia Maffeis}\thanks{Equally contributing}
\affiliation{Department of Physics, Politecnico di Milano, Milan, 20133, Italy}
\author{Lorenzo Chicchi}
\affiliation{Department of Physics and Astronomy, University of Florence, Sesto Fiorentino, 50019, Italy}
\author{Francesco Coghi}
\affiliation{School of Computing and Mathematical Sciences, University of Leicester, Leicester, LE1 7RH, UK}
\affiliation{School of Physics and Astronomy, University of Nottingham, Nottingham, NG7 2RD, UK}
\affiliation{The Alan Turing Institute, London, NW1 2DB, UK}
\author{Duccio Fanelli}
\affiliation{Department of Physics and Astronomy, University of Florence, Sesto Fiorentino, 50019, Italy}
\author{Raffaele Marino}
\affiliation{Department of Physics and Astronomy, University of Florence, Sesto Fiorentino, 50019, Italy}
\affiliation{Butterfly Decisions srl, Via dei Principati 74 - 84122 Salerno, Italy}
\author{Fabrizio Martelli}
\affiliation{Department of Physics and Astronomy, University of Florence, Sesto Fiorentino, 50019, Italy}
\author{Riccardo Paoli}
\affiliation{Computer Science Department, University of Pisa, ISTI-CNR, Pisa, 56127, Italy}
\author{Lorenzo Pattelli}
\affiliation{Istituto Nazionale di Ricerca Metrologica (INRiM), Turin, 10135, Italy}
\author{Lorenzo Spinelli}
\affiliation{Istituto di Fotonica e Nanotecnologie, Consiglio Nazionale delle Ricerche, Milan, 20133, Italy}

\begin{abstract}
The inverse problem of reconstructing optical properties, specifically absorption and scattering coefficients, in layered biological media from time-domain reflectance measurements remains a significant challenge for traditional analytical models. Inverse solvers based on the diffusion equation often struggle with structural heterogeneity, frequently yielding poor accuracy for superficial absorption and deep-layers scattering. In this work, we propose a machine learning framework as an alternative approach to reconstruct the optical properties of a bilayered medium, benchmarking its efficiency and accuracy against  model-based algorithms.
To overcome the intrinsic approximations of diffusion theory and inverse reconstruction,
we generated a robust synthetic dataset of forward DTOF using exact Monte Carlo simulations at multiple source-detector distances.
A machine learning pipeline was then trained on this dataset and validated against state-of-the-art model-based reconstruction methods.
Besides the significant reconstruction speed-up, the machine learning approach achieves higher accuracy than model-based inverse solvers, further providing an estimate of the parameter space dimensionality without requiring any \textit{a priori} information about the number of layers in the investigated geometry.
Further enhancements in the reconstruction accuracy can be expected in future extension of this work, by training the pipeline over multiple DTOF curves from the same medium, in a joint multi-distance reconstruction approach.
\end{abstract}

\maketitle

\section{Introduction}

Light radiative transfer through many natural media, such as biological tissue, is determined by the absorption and scattering coefficients and by the scattering phase function \cite{re2025review}, which constitute the optical properties of the medium. While scattering is related to the physical microstructure of the medium, absorption is linked to the absorption bands of its constituents. These coefficients provide valuable information about the properties of the medium \cite{re2025review}. For instance, in biological tissues, the absorption coefficient provides physiological information about tissue composition through the concentrations of chromophores such as oxy- and deoxyhaemoglobin (thereby reflecting the tissue oxygenation status), as well as water, lipids, and collagen. Conversely, the scattering coefficient provides information about the tissue microstructure. For this reason, the reconstruction/regression of their values has been pursued in many application fields for many years.

These studies are usually implemented in the near-infrared (NIR) therapeutic window from \SIrange[range-units=single]{600}{1100}{\nano\meter}, where light propagation is dominated by scattering with scattering coefficients often exceeding absorption by orders of magnitude. Under these conditions, the detected light is characterized by multiple scattering, and radiative transfer can be simplified by the predictions of the diffusion equation (DE). Consequently, the optical properties of the medium reduce to only the absorption coefficient, $\mua$, and the reduced scattering coefficient, $\musr$, which synthetically incorporate the information of the scattering coefficient and the scattering function. 
Over the past decades, a wide range of techniques have been developed to retrieve these parameters from measurements performed in different domains, including continuous-wave, frequency-domain, and time-domain (TD) systems. This work focuses on the TD due to the fact that other analysis domains can be derived from it.

Since the first attempts to reconstruct $\mua$ and $\musr$ for an unknown medium \cite{Pattersonetal89, Fantini:94, Martellietal97}, their retrieval has traditionally relied on analytical solutions of the DE \cite{Fantini:94, KienlePatterson97, Continietal97, Martellietal97}, combined with nonlinear optimization procedures such as the Levenberg–Marquardt algorithm \cite{Press_etal_88_Numerical}. While these approaches have proven effective in homogeneous media \cite{Martellietal97}, their performance degrades in the presence of structural heterogeneity \cite{Kienleetal98b, KienleGlanzmann99, Martellietal04b}, which is ubiquitous in biological tissues.

Many relevant biological systems --- such as skin, brain, or muscle --- can be approximated as layered media, where a superficial layer overlies a deeper bulk region that may in turn exhibit a further layered structure. Separating the absorption contributions of superficial and deeper tissue layers is crucial in biomedical applications, for example, in distinguishing scalp hemodynamics from cerebral oxygenation in functional near-infrared spectroscopy (fNIRS) \cite{TORRICELLI201428_review_fnirs}.
This has motivated extensive work on the reconstruction of optical properties in bilayer and multi-layer geometries, both \textit{in silico} \cite{Kienleetal98, Kienleetal98b, Martelli:03} and via experimental measurements on phantoms or \textit{in vivo} \cite{Kienleetal98b, KienleGlanzmann99, Martellietal04b}, with the bilayer geometry representing the most commonly used schematization of the medium. 
These previous model-based reconstructions exhibit consistent performance characteristics and limitations, which can be summarized as follows: i) in suitable conditions, high accuracy in retrieving the second layer $\mua$ and the first layer $\musr$, and lower accuracy in retrieving the first layer $\mua$ and the second layer $\musr$; ii) a computational cost dictated by the efficiency of the forward DE solutions; and iii) a sensitivity to the intrinsic approximations of the DE, necessitating the exclusion of early-time data. Notably, most of these foundational studies relied on multi-distance measurement schemes and the non linear regressions were in some cases also tested by reconstructing the first layer thickness of the medium.

Subsequent works \cite{Martelli_Optimal_estimation_two_layers, Martelli_2012_Optimal_Estimation, GARCIA201766_two_three_layers_single_distance, Garcia:18_retrieval_multilayers} have extended these findings to the retrieval of the optical properties of layered media by considering single-distance measurements and by testing Bayesian approaches, such as optimal estimation and the Kalman filter \cite{Martelli_Optimal_estimation_two_layers, Martelli_2012_Optimal_Estimation, GARCIA201766_two_three_layers_single_distance, Garcia:18_retrieval_multilayers}, which --- when \textit{a priori} information on the optical properties is available --- can provide better performance than the Levenberg–Marquardt algorithm \cite{Martelli_2012_Optimal_Estimation, Martelli_Optimal_estimation_two_layers, Garcia:18_retrieval_multilayers}.
\citet{Garcia:18_retrieval_multilayers} considered a four-layer medium and also included the layers thickness in the reconstruction showing that with a limited number of temporal measurements it is possible to map not only ``what'' is inside, but also ``where'' the interfaces between tissues are located. Apart from the different algorithms used to reconstruct the optical properties, as well as the varying number of layers and source–detector configurations considered, these more recent investigations remain consistent with earlier work, in that the main limitations observed in the initial retrieval attempts persist even when the algorithm or measurement setup is modified.

\citet{Geiger:19_P3_three_layers}, although using a P3 approximation model instead of a pure diffusion approach and a three-layer geometry, substantially found results that can be considered in agreement with previous bilayer investigations
since the better reconstructed values found were for the absorption coefficient of the third layer and the reduced scattering coefficient of the upper layer.

In \cite{yang2020space} \citeauthor{yang2020space} propose a novel methodology for retrieving the absolute values of optical absorption and
reduced scattering coefficients in bilayered structures, investigating the ``deep scattering neutrality'' of the results. The accuracy, reliability and linearity of the new method are demonstrated to be better than those obtained by considering a single-domain geometry. This aspect was studied also by \citet{jones2025spectrally} when spatially-resolved reflectance is used for \textit{in vivo} optical characterization of human skin: they showed the potential risks in terms of accuracy if the common simplifying assumption of a single-layer volume is adopted. 

The growing interest in bilayered media is further demonstrated by \cite{krivetskaya2026vivo}, in which the optical properties of both layers were experimentally estimated using the inverse adding-doubling (IAD) method. Unlike time-domain approaches, \citeauthor{krivetskaya2026vivo} exploited spectral information and employed a modified two-stream Kubelka--Munk model to retrieve the optical properties, reporting an accuracy of \SI{86 +- 13}{\percent}.

Also the reconstruction of oxy- and de-oxy haemoglobin concentrations in layered structures was considered. As an example, \citeauthor{sudakou2023two} presented a new method of data analysis for TD NIRS that enables accurate determination of large changes of the absorption coefficient in multiple layers that utilizes changes in moments of DTOF \cite{sudakou2023two}. The method has been tested on a bilayered blood-lipid phantom, demonstrating the accurate determination of $\mua$ and tissue saturation in both compartments.

Finally, a theoretical study using an optimized N-layer diffusion-based solver to reconstruct optical coefficients of bilayered media from TD reflectance has been recently performed in \cite{Rajasekhar_recovery_bilayered_tissues}, substantially confirming the main limitations of the previous results when optical properties are retrieved in such a kind of structures.

In general, previous literature suggests the existence of intrinsic limitations on the achievable accuracy with model-based reconstructions of the optical properties of layered media in the TD, which tends to yield poor accuracy for the absorption of the superficial layer and the scattering of deeper layers. This represents a common thread linking the results reported in all previous studies on the reconstruction of optical properties in layered media.

The previously reviewed works show that reconstructing the optical properties of a bilayer medium presents significant computational challenges for model-based inverse solvers. 
For these reasons, the extraction of optical properties using machine learning algorithms has attracted increasing research interest in recent years. However, most existing approaches are limited to homogeneous media \cite{witteveen2025physics, smith2022deep, nishimura2021determination, panigrahi2019machine, nguyen2021machine, deng2025extraction, balasubramaniam2022tutorial, hokr2021machine}. To the best of our knowledge, only one study has extended this approach to a multilayered medium in the continuous-wave domain~ \cite{tsui2018modelling}. Specifically, this work  considered a four-layer model, with layer thicknesses of the order of micrometers, designed to mimic human skin. The proposed framework estimated the scattering properties of the first three layers, the melanin concentration in the third layer, and the blood volume fraction and oxygen saturation in the fourth layer.

Our study investigates a machine learning framework for extracting the absorption and scattering coefficients of bilayered media from time-domain measurements. The framework is evaluated across very broad ranges of upper-layer thickness (\SIrange[range-phrase={--}, range-units=single]{2}{40}{\milli\meter}), source–detector separation (\SIrange[range-phrase={--}, range-units=single]{10}{30}{\milli\meter}), absorption (\SIrange[range-phrase={--}, range-units=single]{0.0001}{0.1}{\per\milli\meter}) and reduced scattering (\SIrange[range-phrase={--}, range-units=single]{0.5}{1.5}{\per\milli\meter}) coefficients of both layers, encompassing a wide variety of tissues optical properties and, consequently, a broad spectrum of potential applications.  
Its performance is benchmarked against that of conventional model-based reconstruction algorithms in terms of both accuracy and computational efficiency.

Accordingly, our work aims to address whether the limitations observed in model-based reconstructions persist in machine learning approaches or whether a different behavior emerges in this alternative framework. Furthermore, the machine learning framework employed here relies exclusively on exact solutions of the radiative transfer equation (RTE) obtained via Monte Carlo (MC) simulations, resulting in a higher level of accuracy than that achieved by model-based reconstructions on the DE, which are inherently limited by its approximations.

Finally, the primary objective of this study was to assess the fundamental performance of the analyzed methods. To this end, the retrieval was based exclusively on a single time-resolved measurement acquired at one source--detector separation, representing the minimum amount of information that can be extracted from time-domain data. It should be also remarked, that the specific machine learning pipeline here employed (and which leverages a spectral formulation of the optimization process \cite{giambagli2024student, buffoni2022spectral, chicchi2026estimating}), allows for a self-consistent and precise assessment of the relevant free parameters for the problem under scrutiny. No prior knowledge of the underlying geometrical arrangement has to be enforced into the model except for the thickness of the superficial layer to be trained on the available data. Accordingly, this work establishes the fundamental characteristics of machine learning-based time-domain reconstruction and compares them with those of conventional model-based approaches under the most information-limited conditions. Future studies will be required to determine whether these findings remain valid and whether the reconstruction performance can be further improved when inverse procedures exploit a larger amount of information contained in the measurement as for instance the use of multi-distance measurements.
It should also be emphasized that the range of optical and geometrical properties considered in this study is substantially broader than that investigated in previous studies. This difference should be taken into account when comparing our results with those reported in the existing literature.

The work is organized into two main sections: Methods and Results.
The Methods section introduces the notation and describes the generation of the synthetic dataset used in this study. It then presents the classical model-based reconstruction adopted as a reference. Finally, the learning-based reconstruction pipeline --- representing the main innovation of the proposed inverse procedure --- is detailed. The Results section is devoted to presenting and comparing the reconstruction of the optical properties of a bilayer medium obtained using these two independent approaches. Furthermore, in the Conclusion section final considerations about the results achieved are drawn. In particular, the analysis shows that, alongside a marked reconstruction speed-up, the machine learning approach surpasses model-based inverse solvers in accuracy and enables the estimation of the parameter space dimensionality, with the sole input of the superficial layer thickness.

\section{Methods}
\subsection{Problem setup and notation}

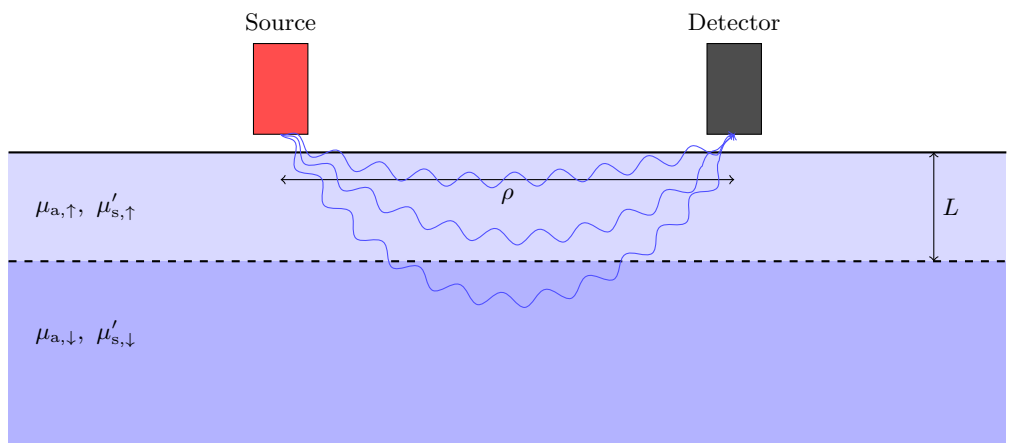
\begin{figure}[htb!]
    \centering
    \begin{tikzpicture}[scale=1.2]

        \fill[blue!15] (-3,0) rectangle (8,-1.2);

        \fill[blue!30] (-3,-1.2) rectangle (8,-3.2);

        \draw[thick] (-3,0) -- (8,0);

        \draw[thick,dashed] (-3,-1.2) -- (8,-1.2);

        \draw[fill=red!70] (-0.3,0.2) rectangle (0.3,1.2);
        \node[above] at (0,1.25) {\small Source};

        \draw[fill=black!70] (4.7,0.2) rectangle (5.3,1.2);
        \node[above] at (5,1.25) {\small Detector};

        \draw[<->] (0,-0.3) -- (5,-0.3)
            node[midway,below] {$\rho$};

        \draw[<->] (7.2,0) -- (7.2,-1.2)
            node[midway,right] {$L$};

        \node[anchor=west] at (-2.8,-0.6) {$\muau,\ \musu$};
        \node[anchor=west] at (-2.8,-2.0) {$\muad,\ \musd$};

        \tikzset{photon/.style={
            blue!70,->,decorate,
            decoration={snake,amplitude=1mm,segment length=6mm}
        }}

        \draw[photon] (0,0.2) .. controls (1.2,-0.4) and (3.0,-0.4) .. (5,0.2);

        \draw[photon] (0,0.2) .. controls (1.4,-1.0) and (3.3,-1.4) .. (5,0.2);

        \draw[photon] (0,0.2) .. controls (1.6,-2.2) and (3.5,-2.0) .. (5,0.2);

    \end{tikzpicture}
    \caption{Schematic representation of a DTOF acquisition in a bilayer medium with a top layer of thickness $L$ and a semi-infinite bottom layer.}
    \label{fig:DiffOpt}
\end{figure}

The paradigmatic case considered here is a bilayer medium with a semi-infinite bottom layer (see Fig.\ref{fig:DiffOpt} for a geometrical schematic in reflectance configuration). In this setup, a monochromatic source is placed on the top layer, emitting a pulse of photons with a delta Dirac instrumental response function (IRF), thereby removing any dependence on the specific IRF shape. The photons then propagate into the bilayer medium, being subject to scattering and absorption events. A detector is placed at a certain (known) distance $\rho$ and counts the number of photons arriving on it, producing a histogram of photon counts in time. These histograms are known as DTOF curves, which we will use as an input of our inverse problem. The effective temporal resolution of a TD system is typically quantified by the Full Width at Half Maximum of the DTOF corresponding to IRF, and is on average \SI{332 +- 250}{\pico\second} \cite{re2025review}. In our paradigmatic case where the IRF is assumed to be a Dirac delta function, the temporal resolution can be determined by the histogram sampling interval, typically $<\SI{100}{\pico\second}$. The thickness $L$ of the top layer is also known in our specific case, in order to focus only on the reconstruction performance of optical properties. This parameter is measurable also in real conditions with ultrasound imaging. The unknowns we will focus on in this work are the absorption and reduced scattering coefficients of the top ($\muau$, $\musu$) and bottom ($\muad$, $\musd$) layers, respectively.

This measurement setup in reflectance was chosen as it mirrors the one employed by most TD systems. It allows for easier probe design and handling, and offers better adaptability to different body regions while maintaining an adequate signal‑to‑noise ratio by controlling the source–detector distance (e.g., head, leg, arm, thorax \cite{re2025review}).
Transmittance is more common to measure breast tissue (which is usually assumed not to be layered), compressed between two parallel glass plates.
 
\subsection{Creation of the bilayer dataset}

A synthetic dataset of time-resolved reflectance curves was generated by means of the GPU-accelerated Monte Carlo software MCX software package, which models photon migration in turbid media by solving light transport without relying on the diffusion approximation \cite{Fang:09}. This choice was motivated by the need to produce a reference dataset as accurate as possible over a broad range of optical properties.

The simulated geometry consisted of a cylindrical bilayer medium of radius $R = \SI{50}{\milli\meter}$ and height $H = \SI{50}{\milli\meter}$, composed of a superficial layer of thickness $L$ overlying a deeper layer of thickness $H-L$. The computational domain was discretized on a three-dimensional voxel grid with spatial resolution of \SI{1}{\milli\meter}. The source was positioned at the center of the upper surface of the cylinder, while photon detection was performed in reflectance geometry at source--detector separations $\rho = \SIlist[list-units=single]{1;2;3}{\centi\meter}$. To exploit the cylindrical symmetry of the problem and improve photon statistics, a ring of detectors was adopted for each source--detector separation.

To efficiently span a large set of absorption values, simulations were carried out according to a white Monte Carlo strategy. In this approach, both layers were initially simulated with zero absorption, while the photon pathlengths due to scattering travelled in each layer were recorded for all detected photons. Absorption was then introduced a posteriori by weighting each detected photon according to the Beer--Lambert law,
\begin{equation}
w = \exp\!\left(-\muau \ell_{\uparrow} - \muad \ell_{\downarrow}\right),
\end{equation}
where \(\ell_{\uparrow}\) and \(\ell_{\downarrow}\) denote the pathlength travelled by the photon in the superficial and deep layer, respectively. This strategy made it possible to generate multiple absorption conditions from the same simulation, substantially reducing the overall computational burden. 

The range of geometrical and optical properties explored in this work was selected to be representative of diffuse optical investigations in biological tissues in the near-infrared window. In particular, the thickness of the superficial layer \(L\) was varied over \SI{2}{\milli\meter} discrete values in the range \SIrange[range-units=single, range-phrase={--}]{2}{40}{\milli\meter}. The absorption coefficients of the upper and lower layers, \(\muau\) and \(\muad\), were sampled over \num{10} values each, spanning the interval \SIrange[range-units=single, range-phrase={--}]{0.0001}{0.5}{\per\milli\meter}. Similarly, the reduced scattering coefficients of the two layers, \(\musu\) and \(\musd\), were sampled over \num{11} values each, covering the interval \SIrange[range-units=single, range-phrase={--}]{0.5}{1.5}{\per\milli\meter}. The full parameter space therefore consisted of \numproduct{10 x 10 x 11 x 11} optical configurations for each combination of \(L\) and \(\rho\).

For each configuration, photon arrival times were collected to construct the corresponding DTOF curve. The temporal axis was discretized into \num{1000} bins of width \SI{10}{\pico\second}, covering a total time window of \SI{10}{\nano\second}. In this study, the instrument response function was assumed to be a Dirac delta function in order to isolate the information content associated with photon propagation in the medium and avoid dependencies on a specific instrumental setup. The number of launched photons was adapted according to the scattering properties of the medium so as to guarantee at least \num{e5} detected photons for each simulated configuration. The single $(\musu,\musd)$ simulation required about 80 seconds using a PC (Intel\textregistered Xeon\textregistered w9-3575X, clock 2.21 GHz, RAM 256 GB) provided with a NVIDIA\textregistered GeForce RTX 5080 as GPU.

After applying the absorption reweighting, each DTOF was optionally normalized and processed before being used in the machine learning pipeline. Overall, the final dataset consisted of a total of \(10 \; \muau \times 10 \; \muad \times 11 \; \musu \times 11 \; \musd \times 3 \; \rho \times 11 \; L = 399300\) unique DTOF curves and was used both to train the spectral autoencoder and to benchmark the subsequent supervised reconstruction models.

A summary of the explored simulation parameters is reported in Table~\ref{tab:dataset_parameters}.

\begin{table}[ht]
\centering
\caption{Summary of the geometrical and optical parameters used to generate the synthetic bilayer dataset.}
\label{tab:dataset_parameters}
\begin{tabular}{lc}
\hline
Parameter & Values \\
\hline
Top-layer thickness \(L\) & \SIlist[list-final-separator={, },list-units=single]{2;4;6;8;10;15;20;25;30;35;40}{\milli\meter} \\
Source--detector separation \(\rho\) & \SIlist[list-final-separator={, },list-units=single]{10;20;30}{\milli\meter} \\
Top-layer absorption \(\muau\) & \SIlist[list-final-separator={, },list-units=single,list-exponents=combine-bracket]{1e-4; 2e-4; 5e-4; 10e-4; 20e-4; 50e-4; 100e-4; 200e-4; 500e-4; 1000e-4}{\per\milli\meter} \\
Bottom-layer absorption \(\muad\) & \SIlist[list-final-separator={, },list-units=single,list-exponents=combine-bracket]{1e-4; 2e-4; 5e-4; 10e-4; 20e-4; 50e-4; 100e-4; 200e-4; 500e-4; 1000e-4}{\per\milli\meter} \\
Top-layer reduced scattering \(\musu\) & \SIrange[range-units=single, range-phrase={--}]{0.5}{1.5}{\per\milli\meter} (step \SI{0.1}{\per\milli\meter}) \\
Bottom-layer reduced scattering \(\musd\) & \SIrange[range-units=single, range-phrase={--}]{0.5}{1.5}{\per\milli\meter} (step \SI{0.1}{\per\milli\meter}) \\
Temporal bin width & \SI{10}{\pico\second} \\
Number of temporal bins & \num[print-unity-mantissa=false]{1e3} \\
Minimum detected photons per DTOF & \num[print-unity-mantissa=false]{1e6} \\
\hline
\end{tabular}
\end{table}

\subsection{Analytical model for reconstruction}
As a benchmark for model-based reconstructions, the analytical model based on the DE for a bilayered medium in reflectance mode described in \cite{Martellietal03,Martellietal04,martelli2022light} was applied for a single source-detector distance and four fitting parameters: the absorption and reduced scattering coefficients of the top and bottom layers. Geometrical parameters as the source-detector distance and the thickness of the top layer were considered known. Indeed, from an experimental standpoint, the first parameter is fixed by the operator, while the second can be estimated through US images. DTOFs are processed from \SIrange{90}{0.5}{\percent} before and after the peak, respectively, considering a Dirac delta as instrument response function. An initial temporal offset of \SI{5}{\pico\second} ps between the simulated DTOF and the IRF, corresponding to half the temporal bin width, was introduced to account for the numerical artifact arising from the discretization of continuous photon arrival times into histogram bins \cite{martelli2022light}.

The accuracy of the retrieved optical parameters depends on a suitable initial estimate, which is then iteratively refined using the Levenberg–Marquardt algorithm within the inversion routine. Inversion stops whether $\chi^2$ tolerance (\num{0.01}) or the maximum number of iterations (\num{100}) are met. The second case occurred to \SI{0.7}{\percent} of entries. In simulations and phantom experiments, the true optical properties are known, but not in vivo. Therefore, to reproduce realistic measurement conditions, a preliminary analytical homogeneous model was applied to the same DTOFs (Formula \ref{eq:td_dtoff}). This approach provides single values of $\mua$ and $\musr$, which do not correspond exactly to either layer but represent an average contribution of both. These effective values were therefore used as initial guesses for the bilayer fit: the estimated absorption was assigned to both $\muau$ and $\muad$, and the estimated reduced scattering coefficient to both $\musu$ and $\musd$.

\begin{figure}[htb!]
    \centering
    \includegraphics[width=0.6\linewidth]{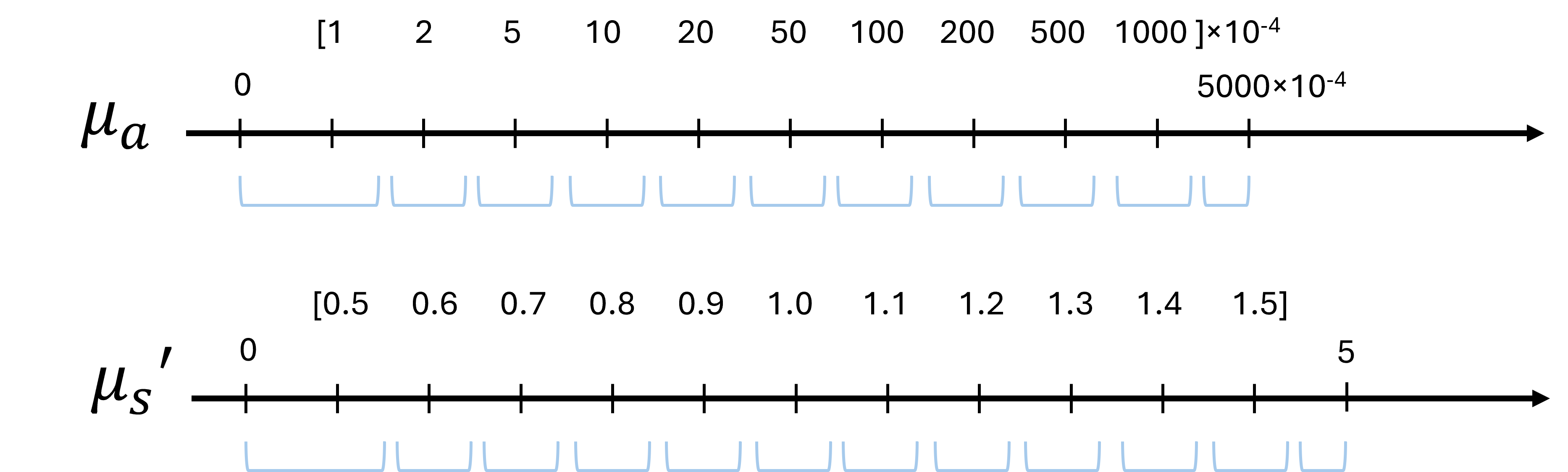}
    \caption{Ranges for accuracy assessment of optical properties, expressed in \si{\per\milli\meter}. Blue parenthesis collect values belonging to the same group. The first and last values, not between square brackets, are used as edges and do not correspond to effective simulations.}
    \label{fig:Ranges}
\end{figure}

After the iterative optimization, a reconstructed value is classified as correct when it falls within the interval defined by the midpoints between the corresponding nominal value and its first neighbors, as depicted with the blue ranges in Figure \ref{fig:Ranges}. For boundary cases, the lower and upper limits were set to \SIlist{0;0.5}{\per\milli\meter} for absorption, and \SIlist{0;5}{\per\milli\meter} for reduced scattering, respectively. Accuracy is computed as the ratio between the number of correct reconstructions and the total number of combinations.

\subsection{Learning-based pipeline for reconstruction}

In order to develop an approach that can be extended to a variety of datasets we first employed a Spectral Autoencoder (SPAE), a variation of a standard deep learning model \cite{rumelhart1985learning, hinton2006reducing} that leverages the spectral reformulation of layers introduced in \cite{giambagli2021machine}.
By the spectral parametrization of the latent dimension, we effectively introduce an eigenvalue for each of the latent variables. These eigenvalues can be learned during training, and through the use of a regularizer, we can automatically discover the correct intrinsic dimension of the input data as demonstrated in Appendix \ref{app:1} for a simpler controlled case.

We first trained a SPAE to extract the minimum number of parameters needed to fully describe our DTOFs. The particular architecture of the SPAE used consisted in 7 fully connected layers: the first and the last of 1000 neurons each, the second and the 6th of 512 neurons each and the 3rd and the 5th of 256 neurons each. The initial dimension of the latent space (4th layer) was set as 15 in order to be much higher than the expected intrinsic dimension of the data. The SPAE was then trained on raw DTOFs for 1000 epochs, using the Adam optimizer with a batch size equal to 512 and learning rate of \num{e-4}. We used L2 regularization on both eigenvector and eigenvalue with relative regularization weights of $\alpha=\num{0.1}$ and $\beta=\num{e-5}$.

After training, we inspected the eigenvalues of the latent space and found that the number of relevant active components in the DTOF data is equal to 4.\\

\begin{figure}[ht]
    \centering
    \includegraphics[width=0.8\linewidth]{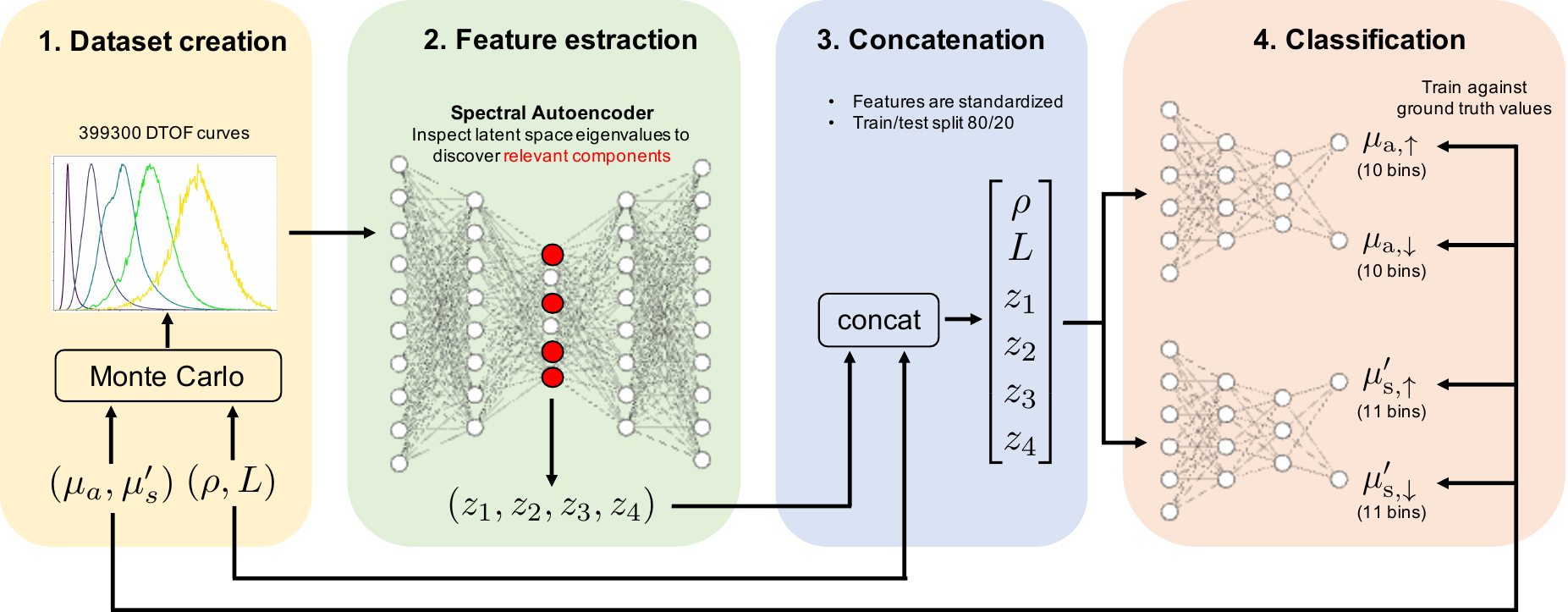}
    \caption{Schematic representation of the learning-based pipeline.}
    \label{fig:scheme}
\end{figure}

After this first training, the encoder is frozen and used as a deterministic feature extractor for the supervised classification stage. The input to the classifier is then a six-dimensional feature vector obtained by concatenating geometry information $(\rho, L)$ and latent coordinates extracted from the SPAE $(z_1,z_2,z_3,z_4)$. The input vector to the classifier thus reads:
\begin{equation}
u = [\,\rho,\;L,\;z_1,\;z_2,\;z_3,\;z_4\,] \in \mathbb{R}^{6}.
\end{equation}

The supervised classification stage predicts discretised versions of the four optical parameters. For each parameter $\theta\in\{ \muau,\muad,\musu,\musd\ \}$, we define a finite set of classes by binning the continuous values into a fixed number of intervals following the exact same procedure described above in Fig.\ref{fig:Ranges}. That means, the absorption parameters are discretised into $B_{\muau} = B_{\muad} = 10$ bins each, while the scattering parameters are discretised into $B_{\musu} = B_{\musd} = 11$ bins each. Each sample is therefore assigned a vector of class labels
\begin{equation}
y_\text{cls} = \bigl(y_{\muau}, y_{\muad}, y_{\musu}, y_{\musd}\bigr),
\end{equation}
with $y_{\theta}\in\{0,\dots,B_{\theta}-1\}$.

Before training the classifier, we standardise each component of $u$ using the mean and standard deviation estimated on the training split only, i.e.\ $u\leftarrow (u-\mu_{\text{train}})/\sigma_{\text{train}}$, and reuse the same transformation at test time. Note that the information used by our classifier is exactly the same as that used in the model-based reconstruction.

The classifier is a multi-head multilayer perceptron (MLP) designed to share a common representation across related targets while allowing separate decision boundaries for each parameter. Rather than using a single four-head network, we train two independent two-head models: one jointly predicts $(\muau,\muad)$ and one jointly predicts $(\musu,\musd)$. This separation keeps each model focused on parameters with similar binning granularity and often improves optimisation stability when different targets have different class cardinalities.

Each two-head MLP takes the vector $u\in\mathbb{R}^{6}$ as input and computes a shared hidden representation through two fully connected layers with ReLU activations,
\begin{equation}
h = \phi\!\left(W_2\,\phi(W_1 u + b_1)+b_2\right),\qquad \phi(\cdot)=\mathrm{ReLU}(\cdot),
\end{equation}
with hidden widths \num{64} and \num{32}. The network then branches into two linear heads that output unnormalised logits for each target,
\begin{equation}
\ell^{(1)} = W^{(1)} h + b^{(1)} \in \mathbb{R}^{B_1},\qquad
\ell^{(2)} = W^{(2)} h + b^{(2)} \in \mathbb{R}^{B_2},
\end{equation}
where $(B_1,B_2) = (10,10)$ for absorption and $(B_1,B_2) = (11,11)$ for scattering. Predicted classes are given by $\hat{y}^{(k)}=\arg\max_j \ell^{(k)}_j$.

We randomly split the dataset into training and test subsets using an $80/20$ partition. The frozen encoder is used to compute $z$ for all samples, and the corresponding feature vectors $u$ and class labels $y_\text{cls}$ are then used to train the MLPs. Both classifiers are optimised with Adam at learning rate \num{5e-3} for up to \num{100} epochs, using a batch size of \num{512}.

For a two-head classifier with logits $(\ell^{(1)},\ell^{(2)})$ and ground-truth labels $(y^{(1)},y^{(2)})$, we minimise the sum of two cross-entropy losses,
\begin{equation}
\mathcal{L}_\text{MLP} = \mathrm{CE}\!\left(\ell^{(1)},y^{(1)}\right) + \mathrm{CE}\!\left(\ell^{(2)},y^{(2)}\right).
\end{equation}
During training, we track both the loss and the top-1 accuracy of each head on the training split.

Performance is quantified using per-parameter top-1 accuracy on the test set, consistently with the analytical benchmark. 

\section{Results}

\begin{figure}[htb!]
    \centering
    \includegraphics[width=1\linewidth]{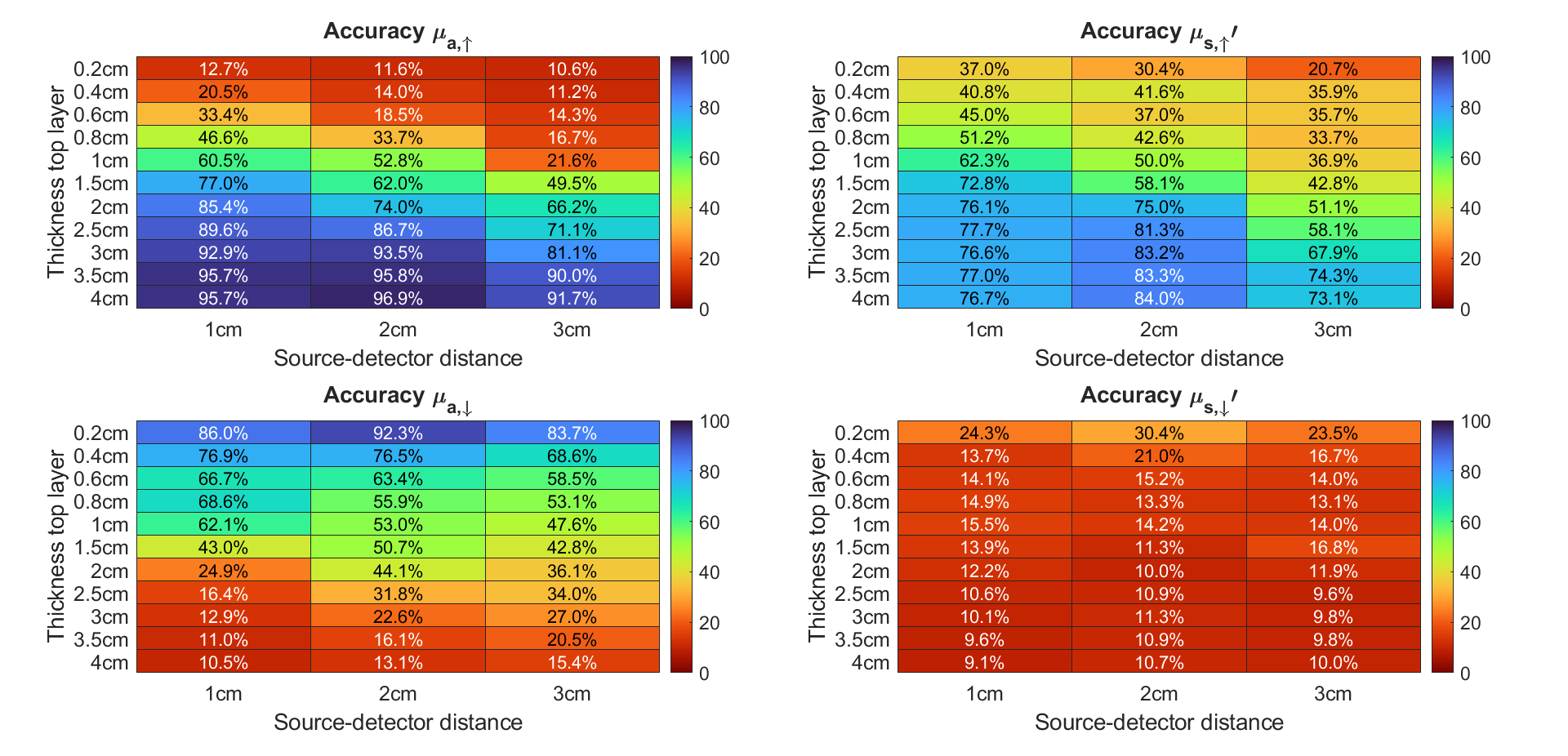}
    \caption{Accuracy results of optical properties retrieved from a model reconstruction based on the analytical solution of the DE for the bilayered medium.}
    \label{fig:analytical}
\end{figure}

Figure \ref{fig:analytical} summarizes the performance of the analytical bilayer model. The four panels correspond to the four fitted parameters and share the same structure: accuracy values are reported for \num{11} top-layer thicknesses (rows) and \num{3} source–detector separations (columns). Each cell aggregates \num{12100} configurations, corresponding to 10 $\muau$ $\times$ 10 $\muad$ $\times$ 11 $\musu$ $\times$ 11 $\musd$ combinations. 

The colormap ranks performance from low accuracy (red) to high accuracy (blue). 

Overall, the reconstructions show moderate average accuracy for $\muau$ (\SIrange[range-phrase={--}]{10.6}{96.9}{\percent}), $\musu$ (\SIrange[range-phrase={--}]{20.7}{84.0}{\percent}), and $\muad$ (\SIrange[range-phrase={--}]{10.5}{92.3}{\percent}), whereas $\musd$ exhibits consistently low accuracy ($\leq$ \SI{30.4}{\percent}).
In general, absorption is retrieved more reliably than scattering.

Performance is strongly influenced by geometry. For a fixed source–detector separation, accuracy improves with increasing top-layer thickness for top-layer parameters, while the opposite trend is observed for the bottom layer, as expected in reflectance geometry. This behavior is consistent with the well-known rule of thumb that the effective photon penetration depth is comparable to the source–detector separation ($L \approx \rho$).

Finally, computational cost is a relevant factor: the full inversion procedure required approximately six days with the same machine used to generate simulations, parallelizing \num{12100} $\times$ 3 $\rho \; \times$ 11 $L$ entries over 40 CPUs.

At first glance, the performance of the model-based reconstruction observed in the present analysis appears to differ from that reported in the studies cited in the Introduction, suggesting a different overall picture. However, before drawing any conclusions, it is important to recognize that the present study explores a substantially broader range of optical and geometric parameters than those considered previously.
This wider parameter space must therefore be taken into account when comparing the results. For example, previous studies reported highly robust retrieval of the absorption coefficient of the second layer compared with that of the first layer, whereas the corresponding estimates obtained in the present work, although still exhibiting good overall accuracy, are somewhat less accurate and do not show a significant improvement over the retrieval accuracy of the first layer.

A plausible explanation is that the expanded parameter space includes more challenging scenarios, such as very large thicknesses of the first layer. Under these conditions, the measurements become significantly less sensitive to the absorption coefficient of the second layer, leading to lower retrieval accuracy. Conversely, the sensitivity to the absorption coefficient of the first layer increases, resulting in improved retrieval accuracy for this parameter. In general, the price one pays in considering such a wide range of the parameter space is that the inclusion of extreme cases can bias the overall assessment of the reconstruction performance compared with the scenarios considered in previous investigations.

Finally, it should also be recalled that several previous studies employed time-resolved datasets acquired at multiple source--detector separations, which inherently provide a larger amount of information.

\begin{figure}[htb!]
    \centering
    \includegraphics[width=1\linewidth]{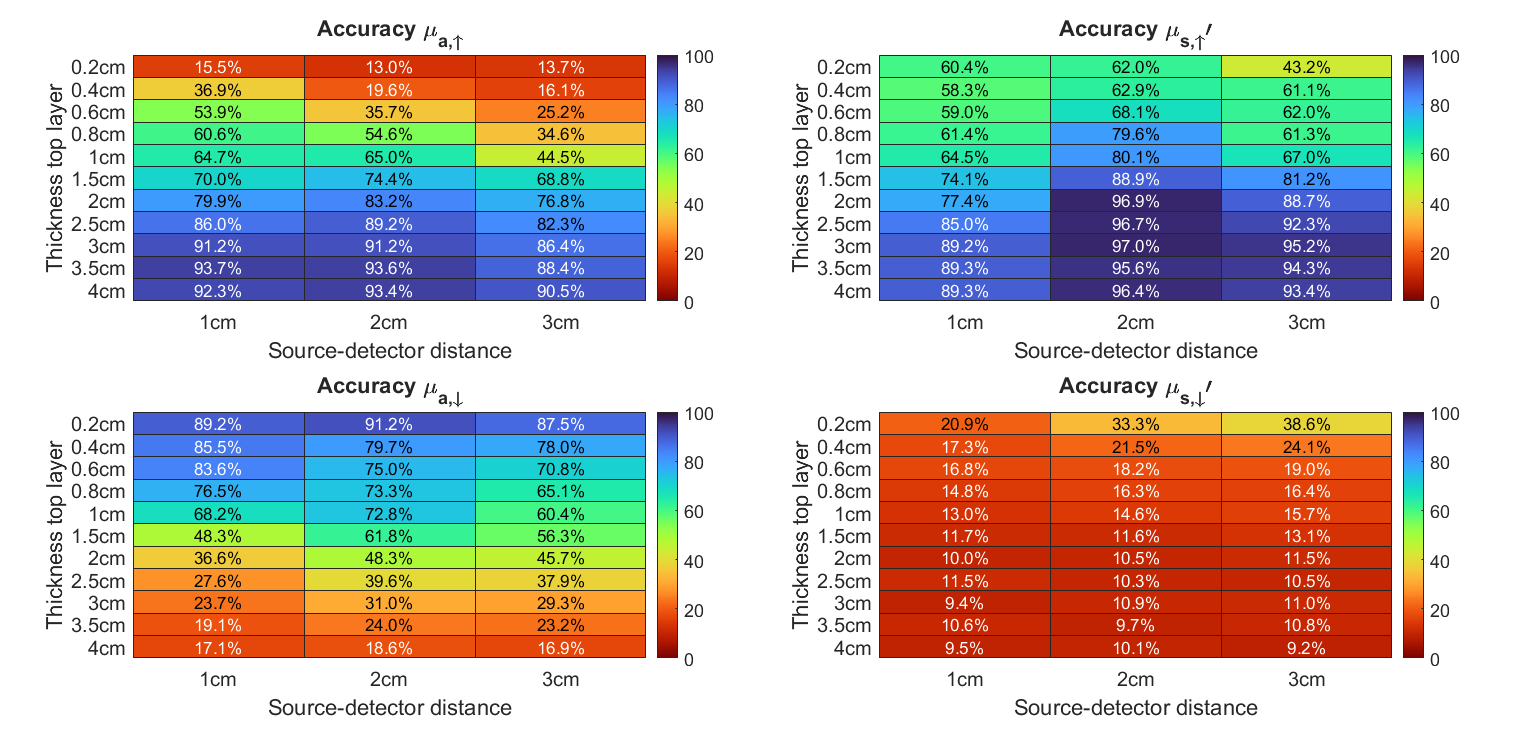}
    \caption{Accuracy results of optical properties retrieved with the ML model for the bilayered medium.}
    \label{fig:learning_based}
\end{figure}

Figure \ref{fig:learning_based} summarizes the performance of the learning-based bilayer model on the test set (i.e.,\ curves never observed at training time). The four panels correspond to the four fitted parameters and share the same structure and meaning as the respective analytical ones in Figure \ref{fig:analytical} above.
The estimates refer to one individual trained model. As one can see the learning-based reconstructions display a consistently higher accuracy than their analytical counterpart, especially regarding the scattering coefficient $\musu$. Still $\musd$ exhibits a very low accuracy ($\leq$ \SI{34}{\percent}) making its reconstruction practically unusable. In general, absorption is retrieved more reliably than scattering.
The computational cost for the learning-based approach was approximately \SI{15}{\min} for training on a single NVIDIA RTX-A5500, while at test-time each new prediction takes just a few milliseconds on the same hardware, being several orders of magnitude faster than the analytical model.\\

\begin{figure}[ht]
    \centering
    \includegraphics[width=1\linewidth]{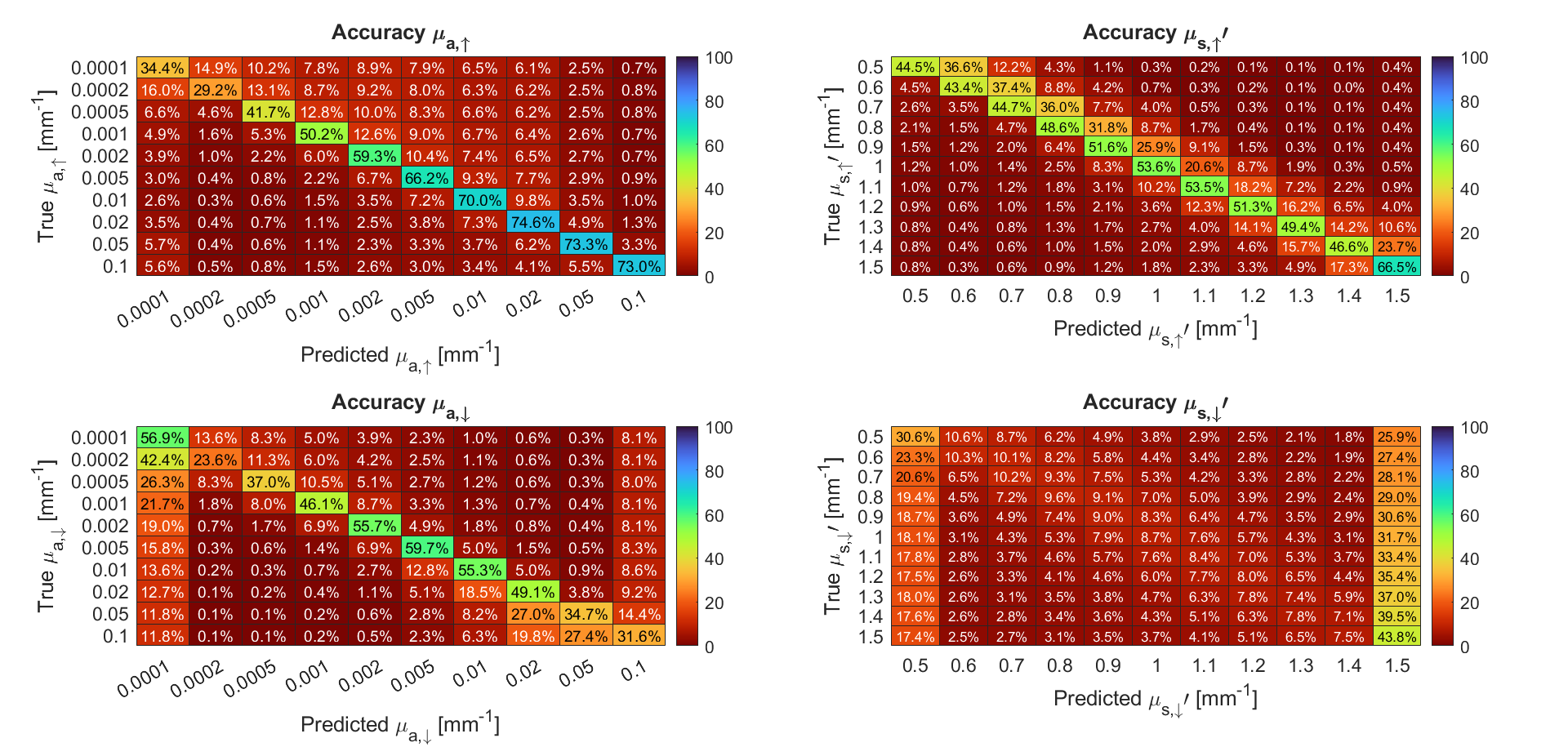}
    \caption{Same data of Figure \ref{fig:analytical} obtained with analytical model, but arranged highlighting performance across optical properties.}
    \label{fig:MuaMusAN}
\end{figure}

\begin{figure}[ht]
    \centering
    \includegraphics[width=1\linewidth]{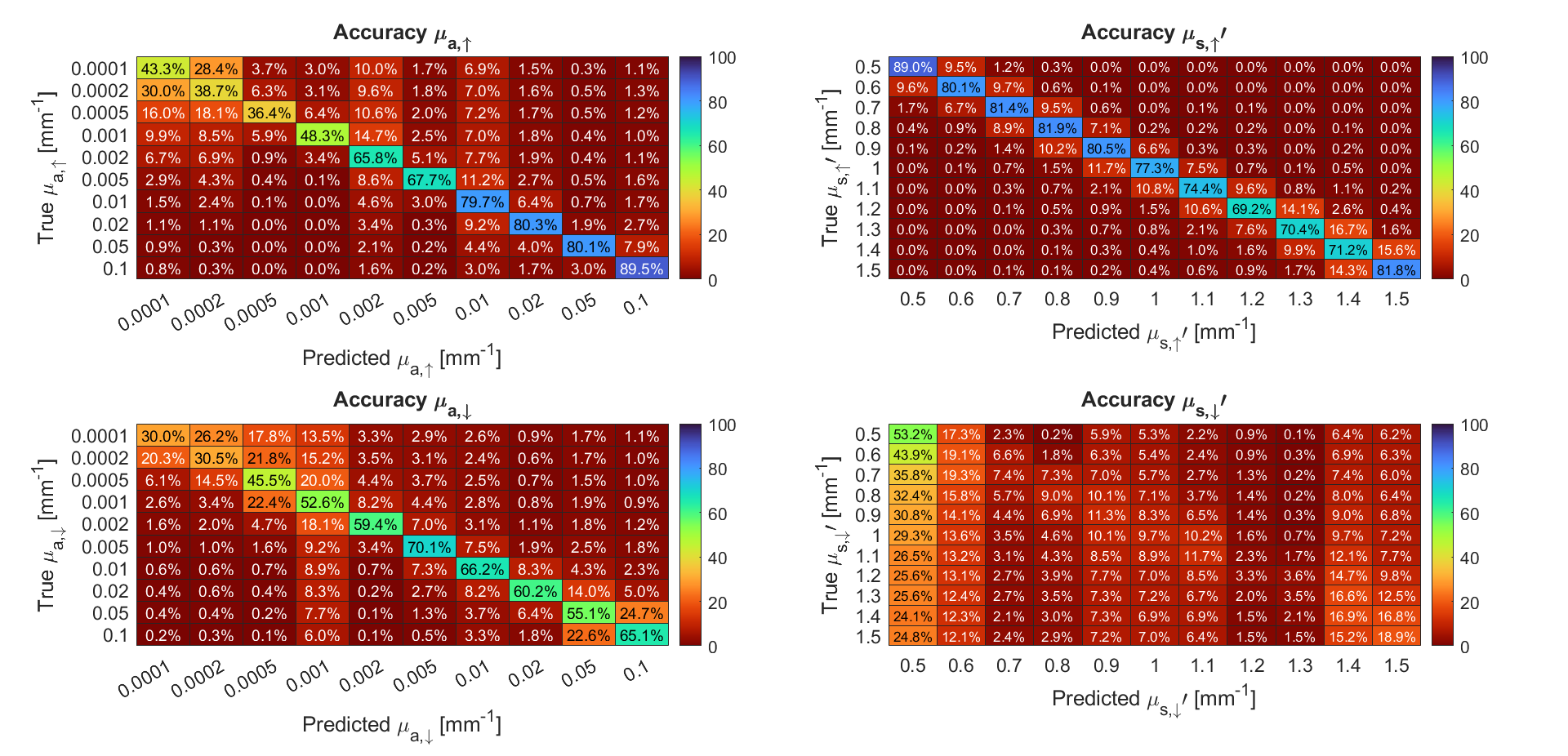}
    \caption{Same data of Figure \ref{fig:learning_based} obtained with ML model, but arranged to highlight performance across optical properties.}
    \label{fig:MuaMusML}
\end{figure}

Figure \ref{fig:analytical} and Figure \ref{fig:learning_based} display results highlighting dependence on geometrical parameters. To understand in which optical range the models perform best, which is particularly relevant for practical use, data have been rearranged, hiding the $\rho$ and $L$ dimensions and rather evidencing the $\muad, \muad, \musu, \musd$ ones. More specifically, using the same dataset, we constructed two groups of four square confusion matrices (one group for the analytical approach - Figure \ref{fig:MuaMusAN}, and one for the ML model - Figure \ref{fig:MuaMusML}), with one matrix corresponding to each optical property of the two layers. In each confusion matrix, rows represent the true values and columns the predicted values (10 classes for $\mua$ and 11 classes for $\musr$). Each cell aggregates results over $3 \; \rho \times 11 \; L$ combinations. With this representation, correct classifications lie along the diagonal. With reference to diagonal elements, the ML method consistently outperforms the analytical approach. On average, the improvement amounts to \SIlist{5.8;8.5}{\percent} for $\mua$ in the first and second layers, respectively, and to \SIlist{27.6;1.0}{\percent} for $\mus^\prime$. Accuracy generally increases with increasing absorption, whereas a slight decrease is observed with increasing scattering. The optical properties retrieved with the ML approach are then reliable ($>\SI{50}{\percent}$) for $\mua>\SI{e-3}{\per\milli\meter}$ and for $\musu \geq \SI{0.5}{\per\milli\meter}$. It is never reliable for $\musd$. In practice this covers most of useful optical ranges for biomedical applications, if we take as a reference the MEDPHOT phantom kit \cite{Pifferi:05} (about \SIrange[range-phrase={--}, range-units=single]{0.005}{0.035}{\per\milli\meter} for absorption and \SIrange[range-phrase={--}, range-units=single]{0.5}{2.0}{\per\milli\meter} for scattering).

Let us now consider off-diagonal elements to understand the errors made by both methods. Table \ref{tab:error} measures errors in terms of class shifts. This metric is particularly appropriate in the present context because it provides a common basis for comparing the two approaches. Indeed, it should be recalled that the analytical (AN) method estimates the optical properties as continuous values, whereas the ML method performs discrete classification, since it can only produce outputs corresponding to the predefined set of input classes.

For each confusion matrix, every element is assigned a class shift, $\Delta C_{row,col}$, defined as its distance from the diagonal along the corresponding row, with the diagonal elements having zero shift, i.e., $\Delta C_{row,row} = 0$. Then, the mean class shift is computed as: 
\begin{equation}B_{row} = \frac{\sum_{col=1}^{N_{col}} \Delta C_{row,col} \times Accuracy_{row,col}}{\sum_{col=1}^{N_{col}} Accuracy_{row,col}},\label{eq:bar}
\end{equation}
where $N_{col} = 10$ for absorption and 11 for scattering. According to this metric, the ML model consistently outperforms the AN approach for all optical properties once again except for $\musd > \SI{1.1}{\per\milli\meter}$. The average class shift across the full parameter range (i.e., rows) decreases from \numrange{1.2}{0.85} for $\muau$, from \numrange{1.56}{0.93} for $\muad$, and from \numrange{0.83}{0.28} for $\musu$. For $\musd$, the improvement is marginal, with the average shift changing from \numrange{3.53}{3.39}.
This behaviour is consistent with the corresponding confusion matrices, which reveal a strong tendency for $\musd$ predictions to accumulate at the extreme classes of the range. 
In addition, we conducted some analysis to rule out spurious effects in these results due to random chance and dependence on initial conditions. This analysis, which can be found in Appendix \ref{app2}, validates the above results, accounting for the variability intrinsic to the training of the ML model, with the analytical benchmark performing better only when initialized with the correct guess (upper bound). The comparison with the analytical method under ideal initialization shows that ML provides consistently more accurate optical property estimates under realistic conditions where no prior knowledge of the parameters is available. Although the analytical method can outperform ML when initialized with the exact solution, this best-case scenario is generally unattainable \textit{in vivo}, confirming the practical advantage and robustness of the ML approach.\\

\begin{table}[ht]
\caption{Error overview of AN and ML reconstruction in terms of $B_{row}$ (Equation \ref{eq:bar}, lower is better). The left-side table refers to absorption results. The right-side table refers to scattering results.}
\begin{tabular}{lllllllllll|ll|ll}
\multicolumn{1}{l|}{}                                & \multicolumn{2}{c|}{\textbf{$\muau$}}              & \multicolumn{2}{c}{\textbf{$\muad$}} &  &  &  &  &  & \textbf{}                                & \multicolumn{2}{c|}{\textbf{$\musu$}} & \multicolumn{2}{c}{\textbf{$\musd$}} \\ \cline{2-5} \cline{12-15} 
\multicolumn{1}{l|}{{[}mm$^{-1}$]} & \textbf{AN}   & \multicolumn{1}{l|}{\textbf{ML}}   & \textbf{AN}       & \textbf{ML}      &  &  &  &  &  & {{[}mm$^{-1}$]} & \textbf{AN}       & \textbf{ML}       & \textbf{AN}       & \textbf{ML}      \\ \cline{1-5} \cline{11-15} 
\multicolumn{1}{l|}{\textbf{0.0001}}                 & 2.43          & \multicolumn{1}{l|}{1.58}          & 1.58              & 1.76             &  &  &  &  &  & \textbf{0.5}                             & 0.88              & 0.13              & 4.12              & 2.13             \\
\multicolumn{1}{l|}{\textbf{0.0002}}                 & 1.98          & \multicolumn{1}{l|}{1.37}          & 1.64              & 1.30             &  &  &  &  &  & \textbf{0.6}                             & 0.82              & 0.21              & 3.97              & 2.26             \\
\multicolumn{1}{l|}{\textbf{0.0005}}                 & 1.53          & \multicolumn{1}{l|}{1.32}          & 1.55              & 0.97             &  &  &  &  &  & \textbf{0.7}                             & 0.80              & 0.22              & 3.78              & 2.47             \\
\multicolumn{1}{l|}{\textbf{0.001}}                  & 1.17          & \multicolumn{1}{l|}{1.09}          & 1.49              & 0.80             &  &  &  &  &  & \textbf{0.8}                             & 0.74              & 0.22              & 3.60              & 2.70             \\
\multicolumn{1}{l|}{\textbf{0.002}}                  & 0.88          & \multicolumn{1}{l|}{0.86}          & 1.42              & 0.70             &  &  &  &  &  & \textbf{0.9}                             & 0.73              & 0.24              & 3.50              & 2.94             \\
\multicolumn{1}{l|}{\textbf{0.005}}                  & 0.67          & \multicolumn{1}{l|}{0.66}          & 1.35              & 0.61             &  &  &  &  &  & \textbf{1}                               & 0.75              & 0.28              & 3.38              & 3.20             \\
\multicolumn{1}{l|}{\textbf{0.01}}                   & 0.58          & \multicolumn{1}{l|}{0.46}          & 1.36              & 0.69             &  &  &  &  &  & \textbf{1.1}                             & 0.79              & 0.33              & 3.31              & 3.49             \\
\multicolumn{1}{l|}{\textbf{0.02}}                   & 0.64          & \multicolumn{1}{l|}{0.42}          & 1.46              & 0.80             &  &  &  &  &  & \textbf{1.2}                             & 0.88              & 0.40              & 3.26              & 3.95             \\
\multicolumn{1}{l|}{\textbf{0.05}}                   & 0.94          & \multicolumn{1}{l|}{0.40}          & 1.66              & 0.88             &  &  &  &  &  & \textbf{1.3}                             & 0.93              & 0.38              & 3.27              & 4.33             \\
\multicolumn{1}{l|}{\textbf{0.1}}                    & 1.18          & \multicolumn{1}{l|}{0.33}          & 2.06              & 0.79             &  &  &  &  &  & \textbf{1.4}                             & 0.94              & 0.35              & 3.29              & 4.55             \\ \cline{1-5}
\multicolumn{1}{l|}{\textbf{Average}}                & \textbf{1.20} & \multicolumn{1}{l|}{\textbf{0.85}} & \textbf{1.56}     & \textbf{0.93}    &  &  &  &  &  & \textbf{1.5}                             & 0.85              & 0.27              & 3.32              & 5.23             \\ \cline{11-15} 
                                                     &               &                                    &                   &                  &  &  &  &  &  & \textbf{Average}                         & \textbf{0.83}     & \textbf{0.28}     & \textbf{3.53}     & \textbf{3.39}   
\end{tabular}\label{tab:error}
\end{table}

To conclude, we observed how the proposed ML pipeline achieves a higher classification accuracy across a wide range of optical and geometrical parameters, combined with a lower error dispersion, and using a fraction of the computational cost of the analytical benchmark.

\section{Conclusion}

In this work, we investigated a learning-based framework for the reconstruction of optical properties in bilayer media from time-resolved reflectance measurements. The proposed approach combines a spectral autoencoder for data-driven dimensionality reduction with supervised classification trained to retrieve the absorption and reduced scattering coefficients of the superficial and deep layers. Its performance was benchmarked against a classical analytical inversion procedure based on the diffusion equation and Levenberg--Marquardt optimization.

The results show that the learning-based approach provides a clear advantage over the analytical benchmark in terms of both computational efficiency and reconstruction accuracy. Once trained, the model performs inference in a few milliseconds, thereby opening the possibility of near real-time retrieval of optical parameters during measurements. This represents a substantial improvement over the analytical inversion, indicating that data-driven reconstructions trained on Monte Carlo simulations can partially overcome limitations associated with diffusion-based inverse solvers, especially when no reliable \textit{a priori} initialization is available. 

We also note that the reconstruction of the reduced scattering coefficient of the deeper layer remains unreliable, even within the learning-based framework. This indicates that the poor sensitivity to deep-layer scattering is not merely a consequence of the analytical inversion method or of diffusion-theory approximations, but may reflect a more fundamental limitation of single-distance time-resolved reflectance measurements in bilayer geometries.

A further strength of the proposed framework is its ability to estimate the intrinsic dimensionality of the DTOF dataset through the spectral autoencoder. The identification of four relevant latent variables is consistent with the four optical parameters governing the light propagation in bilayered diffuse media, suggesting that this strategy may provide useful information on the effective number of degrees of freedom required to describe a given measurement configuration. In perspective, this feature could be exploited not only for parameter reconstruction, but also for model selection and for the automatic identification of the complexity of layered structures.

The rapid spread of machine learning–based reconstruction methods raises the important question of reconsidering classical model-based approaches, in order to determine which methodology is best suited to the various problems being addressed. Furthermore, the scientific community is called upon to deepen its understanding of these approaches, identifying their respective strengths and weaknesses and, ultimately, exploring the possibility of a synthesis between them. Indeed, recent literature reflects this need, with studies beginning to compare traditional and machine learning reconstruction techniques and to review the state-of-the-art in light of recent advances in artificial intelligence, as evidenced in \cite{Hussainetal_AI_and_Classical_skin, Rossberg16032026_AI_NIRS_reflectance}. 
Several directions of development remain open. The use of multi-distance measurements should be systematically investigated, as they may provide complementary depth sensitivity and improve the retrieval of deep-layer scattering. The framework should then be extended to more realistic acquisition conditions, including finite instrument response functions, experimental noise, detector effects, and uncertainty in the layer thickness, in order to ultimately move to \textit{in vivo} datasets. 

To conclude, given the result obtained in this work, there is the possibility that in the future hybrid approaches combining the physical interpretability of model-based inversion with the flexibility and speed of machine learning methods may offer a promising route toward accurate, efficient, and clinically applicable reconstruction of optical properties in heterogeneous biological tissues.

\section*{Data and Code Availability}

The code used to implement the learning-based reconstruction is publicly available on \href{https://github.com/F-Coghi/mlp_dtof}{GitHub}. The dataset generated is similarly available on \href{https://doi.org/10.5281/zenodo.22141275}{Zenodo}.

\section*{Acknowledgements}

FC is supported by a Leverhulme Early Career Fellowship No.\ ECF-2025-482.

\bibliography{bibliography}

\appendix
\section{Reconstruction on simpler DTOFs}
\label{app:1}
In order to validate the capacity of the SPAE model to reconstruct the minimum necessary dimensions to fully describe a given dataset and at the same time to confirm the ability of the process to reconstruct the optical parameters we tested the SPAE on two similar controlled cases.

\begin{figure}[htb!]
    \centering
    \begin{tikzpicture}[scale=1.2]

        \fill[blue!20] (-3,0) rectangle (8,-2.0);

        \draw[thick] (-3,0) -- (8,0);

        \draw[fill=red!70] (-0.3,0.2) rectangle (0.3,1.2);
        \node[above] at (0,1.25) {\small Source};

        \draw[fill=black!70] (4.7,0.2) rectangle (5.3,1.2);
        \node[above] at (5,1.25) {\small Detector};

        \draw[<->] (0,-0.3) -- (5,-0.3)
            node[midway,below] {$\rho$};

        \tikzset{photon/.style={
            blue!70,->,decorate,
            decoration={snake,amplitude=1mm,segment length=6mm}
        }}

        \draw[photon] (0,0.2) .. controls (1.2,-0.6) and (3.0,-0.6) .. (5,0.2);
        \draw[photon] (0,0.2) .. controls (1.4,-1.2) and (3.3,-1.1) .. (5,0.2);
        \draw[photon] (0,0.2) .. controls (1.6,-1.8) and (3.5,-1.7) .. (5,0.2);

    \end{tikzpicture}
    \caption{Schematic representation of a DTOF acquisition in an homogeneous, semi-infinite medium}
    \label{fig:homogeneous}
\end{figure}
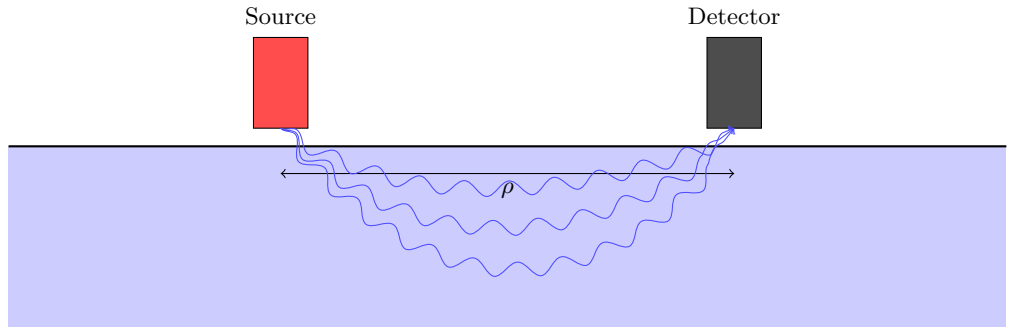

For both these tests we consider an homogeneous, semi-infinite medium (see Fig.\ref{fig:homogeneous}). In this particular setup the time-resolved reflectance at distance $\rho$ is given by the analytical Formula \ref{eq:td_dtoff}, from which we generated a dataset consisting of \num{10000} DTOFs, each one subdivided into \num{1000} bins of \SI{5}{\pico\second}.

For the DE solution for the reflectance of a semi-infinite medium, we used the well-known expression derived using the extrapolated boundary condition \cite{Continietal97,re2025review}: 
\begin{equation}
R(\rho, t) = \frac{1}{2 (4 \pi D v)^{3/2} t^{5/2}} 
\exp\Big(-\mua v t - \frac{\rho^2}{4 D v t}\Big) 
\Big[z_0^+ \exp\Big(-\frac{(z_0^+)^2}{4 D v t}\Big) - z_0^- \exp\Big(-\frac{(z_0^-)^2}{4 D v t}\Big)\Big],
\label{eq:td_dtoff}
\end{equation}
with
\begin{equation}
\left\{
\begin{array}{l}
z_0^+ = z_s \\
z_0^- = - 2z_e - z_s,
\end{array}
\right.
\label{3.12}
\end{equation}
where $z_e = 2 AD$ and $D=1/(3\musr)$ diffusion coefficient, $A$ factor depending the on Fresnel reflections \cite{Continietal97,re2025review}.

For the first test we obtained our dataset selecting the absorption coefficient $\mua$ from a uniform distribution between \SIlist{0.0; 100.0}{\per\centi\meter}.

Such dataset was than given in input into a SPAE with 7 fully connected layers built as following: the first and the last of \num{1000} neurons each, the second and the 6th of \num{512} neurons each, the 2nd and the 4th of 256 neurons each and the latent layer of 10 neurons.

We trained such SPAE for \num{1000} epochs with batch size equal to \num{512} and learning rate \num{e-3}. The regularization on the eigenvalues and the eigenvectors of type L2. The regularization parameters equal to $\alpha = \num{0.1}$ for the eigenvalues and $\beta = \num{e-5}$ for the eigenvectors.

\begin{figure}[htb!]
    \centering
    \includegraphics[width=0.7\linewidth]{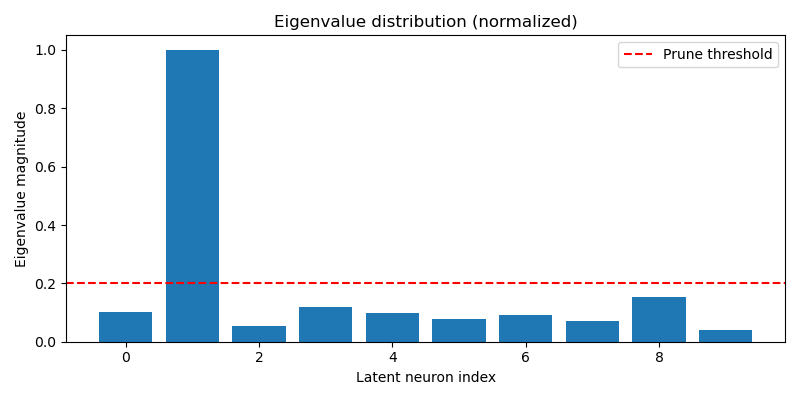}
    \caption{Absolute values of the eigenvectors relative to the neurons of the latent space}
    \label{fig:eig_mua}
\end{figure}

The results of the training show that the relevant eigenvalue is only one (see Fig.\ref{fig:eig_mua}), as a first demonstration of the ability of the SPAE to find the minimum number of parameters needed to reconstruct the given dataset.

In order to further test the SPAE model we considered a more complex case where the input was a DTOF curve from a bilayer medium plus a scalar value representing the source-detector distance (SDD). Theoretically, we would expect this problem to be well represented by 4 dimensions for the DTOF plus 1 for the SDD.

\begin{figure}[htb!]
     \centering
     \includegraphics[width=0.7\linewidth]{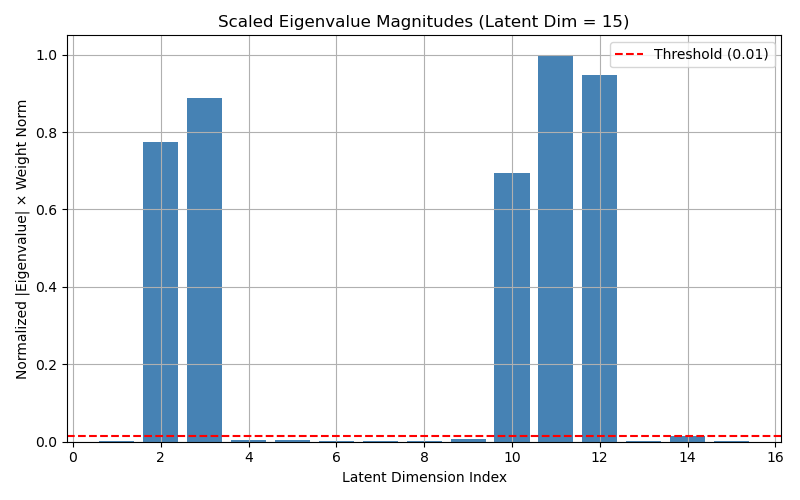}
     \caption{Eigenvalues relative to the latent space of the SPAE.}
     \label{fig:refl_eig_5}
 \end{figure}

Training a SPAE with a latent dimension much greater than the expected one (in this case the latent dimension was chosen to be 15), the test showed clearly that the number of relevant ($> \num{0.01}$) eigenvalues relative to the neurons in the latent dimension is equal to 5 as it can be observed in Figure \ref{fig:refl_eig_5}. It then makes sense to hypothesize that the 5 relevant dimensions of the latent space of the SPAE are linked to some combination of the optical parameters of the medium and the SDD.

\begin{figure}[htb!]
    \centering
    \includegraphics[width=0.7\linewidth]{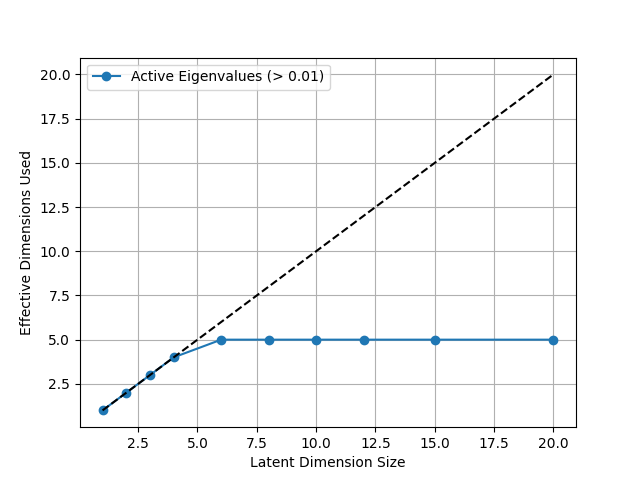}
    \caption{Number of relevant eigenvalues with respect to the dimension of the latent space of the SPAE}
    \label{fig:refl_num_eig}
\end{figure}

To further test the robustness of the intrinsic dimension found, we ran different SPAEs in sequence with different latent dimensions, recording the number of relevant dimensions that emerged from each of  the runs. The results in Figure \ref{fig:refl_num_eig} show that when the latent dimension of the SPAE reaches the "correct" number of parameters that well describes the dataset (in our case 5), the relevant dimensions used plateau and remain unchanged for the bigger models.

\section{Initial condition analysis}
\label{app2}

\begin{figure}[htb!]
    \centering
    \includegraphics[width=1\linewidth]{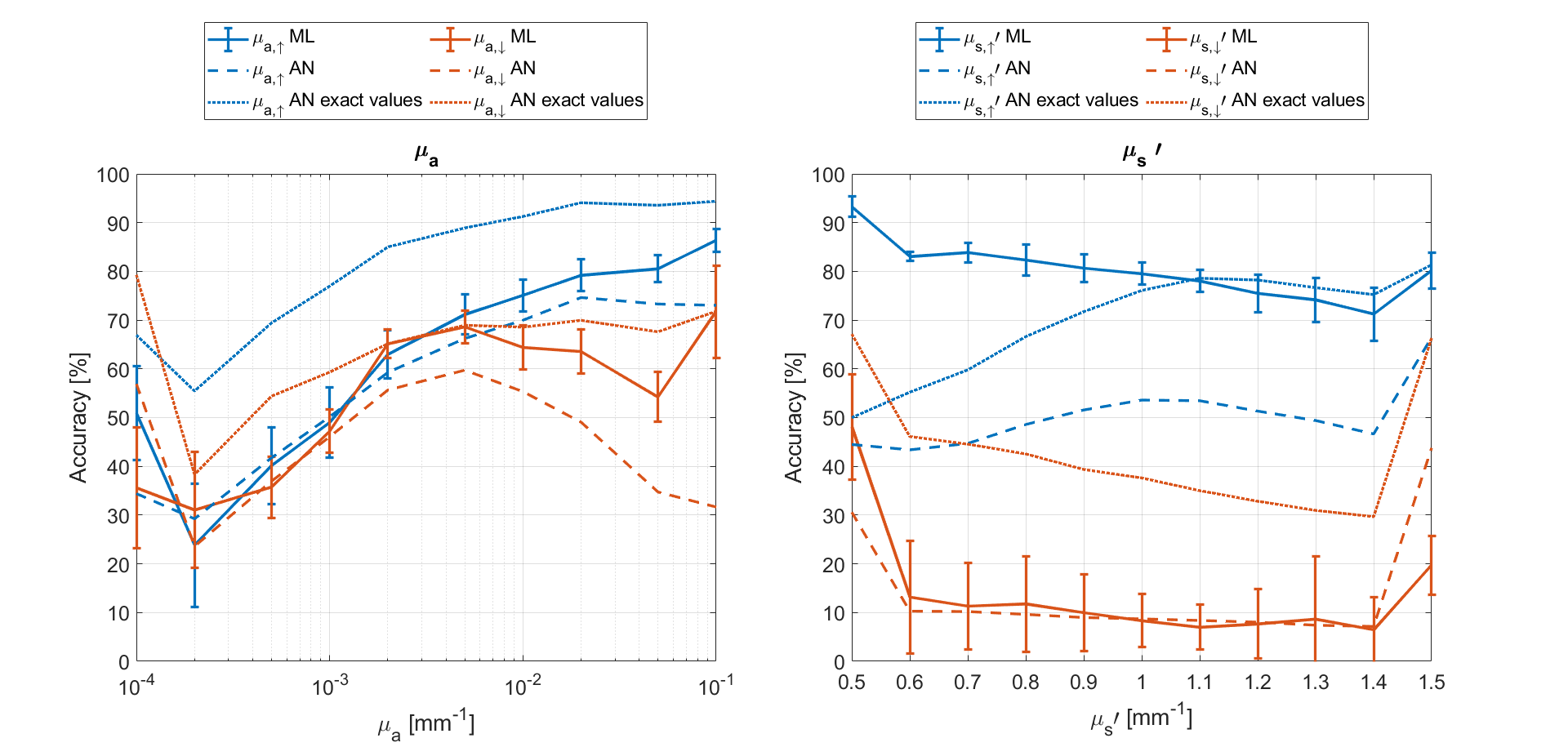}
    \caption{Representation of upper bounds in accuracy achievable by ML (through error bars) and AN (through exact initial guesses) methods.}
    \label{fig:UpperBounds}
\end{figure}

In Figure \ref{fig:UpperBounds}, we present a comparison of the upper bounds in accuracy achievable by both methods, in order to evaluate the robustness of the improvements observed in the main text for the ML approach. The analysis is restricted to correctly classified cases. For the ML model, variability is estimated by repeating the training and validation on the test set 10 times on different random train/test splits and computing the corresponding mean and standard deviation. For the AN approach, this procedure cannot be applied, since the fitting process is deterministic and identical initial guesses yield identical results.
To obtain a meaningful comparison, we therefore repeat the AN fitting in the best-case scenario to obtain an upper bound. Specifically, we initialize the inversion procedure with the exact optical property values. In this configuration, the inputs coincide with the expected outputs, thus representing an ideal reconstruction scenario.
From the figure, it is evident that for $\mua > \SI{0.001}{\per\milli\meter}$ and for all values of $\musu$, the ML error bars do not overlap with the corresponding AN curves, confirming the stronger performance of the ML method. Conversely, as expected, ML performs less accurately than AN when the latter is initialized with exact parameter values. The only exception is $\musu$, where the two methods show comparable performance, with ML outperforming AN for $\musu < \SI{1.1}{\per\milli\meter}$.
It should be emphasized that this best-case scenario is feasible in simulations and measurements on tissue-mimicking phantoms, but is unrealistic for \textit{in vivo} applications, which represent the true target. Overall, these results indicate that ML provides consistently more accurate estimates of optical properties when no ideal initial guess is available.

\end{document}